\documentclass[twocolumn, switch]{article} 

\usepackage{preprint}
\usepackage{algorithm}
\usepackage{algorithmic}
\usepackage{amsmath, amsthm, amssymb, amsfonts}

\usepackage[numbers,square]{natbib}
\usepackage[utf8]{inputenc}	
\usepackage[T1]{fontenc}	
\usepackage{xcolor}		
\usepackage[colorlinks = true,
            linkcolor = purple,
            urlcolor  = blue,
            citecolor = cyan,
            anchorcolor = black]{hyperref}	
\usepackage{booktabs} 		
\usepackage{nicefrac}		
\usepackage{microtype}		
\usepackage{lineno}		
\usepackage{float}			
\usepackage{multirow}
\usepackage{float}  
\usepackage{tabularx}  
\usepackage{preprint}
\usepackage{enumitem}

\usepackage{lipsum}		

\usepackage{newfloat}
\DeclareFloatingEnvironment[name={Supplementary Figure}]{suppfigure}
\usepackage{sidecap}
\sidecaptionvpos{figure}{c}

\usepackage{titlesec}
\titlespacing\section{0pt}{12pt plus 3pt minus 3pt}{1pt plus 1pt minus 1pt}
\titlespacing\subsection{0pt}{10pt plus 3pt minus 3pt}{1pt plus 1pt minus 1pt}
\titlespacing\subsubsection{0pt}{8pt plus 3pt minus 3pt}{1pt plus 1pt minus 1pt}

\title{LDRFusion: A LiDAR-Dominant multimodal refinement
framework for 3D object detection}

\usepackage{eso-pic}
\usepackage{tikz}
\usepackage{xcolor}
\PassOptionsToPackage{colorlinks=true,linkcolor=gray,urlcolor=gray}{hyperref}

\usepackage{titling}
\usepackage{orcidlink}
\usepackage{footmisc}
\newcommand{\Author}[3]{
  \textbf{#1}\textsuperscript{#2}\ %
}

\author{
  \Author{Jijun Wang}{1}{0000-0000-0000-0000} \and
  \Author{Yan Wu}{1}{0000-0000-0000-0000}\and
  \Author{Yujian Mo}{1}{0000-0000-0000-0000}\and
  \Author{Junqiao Zhao}{1}{0000-0000-0000-0000} \and
  \Author{Jun Yan}{2}{0000-0000-0000-0000} \and
  \Author{Yinghao Hu}{1}{0000-0000-0000-0000}
}

\date{%
  \textsuperscript{1}School of Computer Science and Technology, Tongji University, Shanghai 201804, China\\
  \textsuperscript{2}School of Electronics and Information Engineering, Tongji University, Shanghai 201804, China\\[1em]
  \footnotesize \textbf{Corresponding author:} Yan Wu\\
}

\begin{document}

\twocolumn[ 
  \begin{@twocolumnfalse} 

\maketitle
\thispagestyle{empty}

\begin{abstract}
Existing LiDAR-Camera fusion methods have achieved strong results in 3D object detection. To address the sparsity of point clouds, previous approaches typically construct spatial pseudo point clouds via depth completion as auxiliary input and adopts a proposal-refinement framework to generate detection results. However, introducing pseudo points inevitably brings noise, potentially resulting in inaccurate predictions. Considering the differing roles and reliability levels of each modality, we propose LDRFusion, a novel Lidar-dominant two-stage refinement framework for multi-sensor fusion. The first stage soley relies on LiDAR to produce accurately localized proposals, followed by a second stage where pseudo point clouds are incorporated to detect challenging instances. The instance-level results from both stages are subsequently merged. To further enhance the representation of local structures in pseudo point clouds, we present a hierarchical pseudo point residual encoding module, which encodes neighborhood sets using both feature and positional residuals. Experiments on the KITTI dataset demonstrate that our framework consistently achieves strong performance across multiple categories and difficulty levels.
\end{abstract}
\vspace{0.35cm}

  \end{@twocolumnfalse} 
] 



\section{Introduction}

In the field of autonomous driving and advanced driving assistance systems (ADAS)~\cite{MO2022626, qian20223d, mao20233d, yan2025adversarial}, 3D object detection provides critical spatial information, such as the size, orientation, and precise location of objects in the real world. This capability is essential for downstream tasks like trajectory prediction, collision avoidance, and decision-making in complex driving scenarios.

Various sensors can serve as inputs for the 3D detection task, including LiDAR, RGB cameras, millimeter-wave radar etc. Among them, LiDAR stands out for its ability to provide precise positional coordinates. Therefore, many methods~\cite{qi2017pointnet, zhou2018voxelnet, yan2018second, lang2019pointpillars, shi2019pointrcnn, deng2021voxel, yin2021center, yang2022graph, mo2025enhancinglidarpointfeatures} been developed to process point cloud data and achieve effective object recognition. Considering the spatial structural characteristics of point clouds, existing methods often employ voxel-based data processing methods and utilize sparse convolution for feature extraction. To further enhance detection capabilities, some methods adopt a two-stage detection framework, leveraging RoI pooling~\cite{deng2021voxel} or graph neural networks~\cite{yang2022graph} to refine proposals generated by the Region Proposal Network (RPN). 

\begin{figure*}[htpb]
\centering
\includegraphics[scale=0.45]{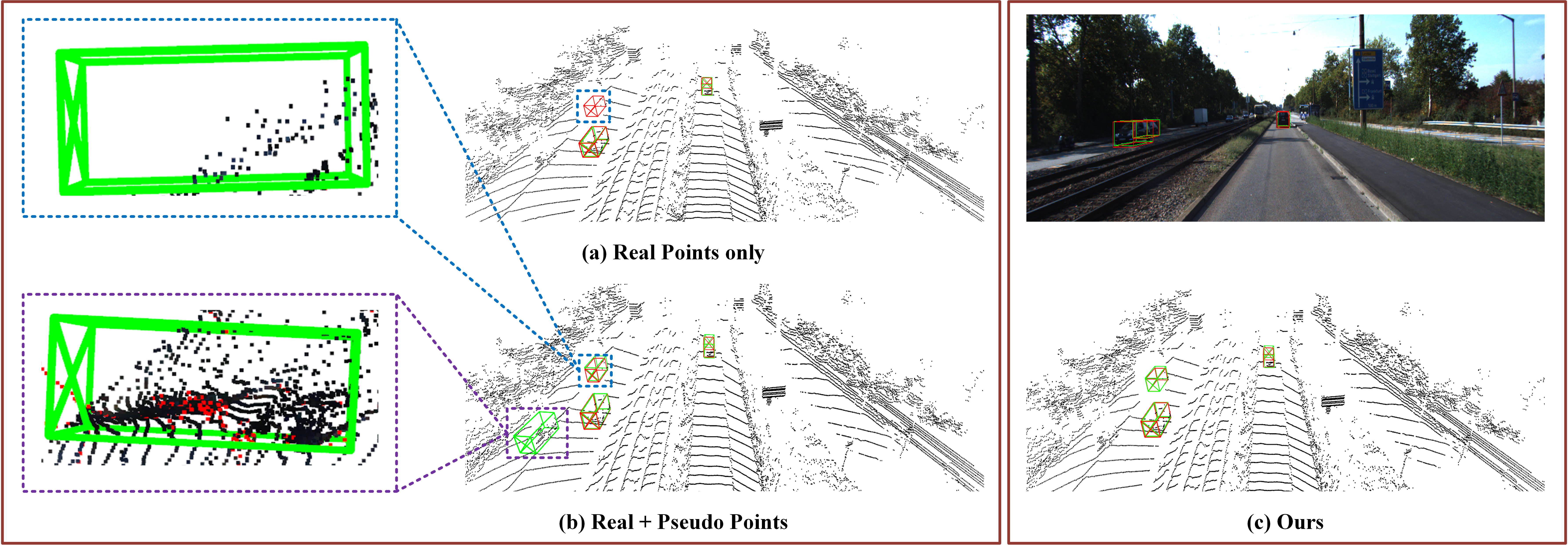}
\caption{\textbf{Visualization of detection results from different models:} (a) using real point clouds, (b) using real and pseudo point clouds, and (c) using the proposed refinement strategy (ours). Ground truths are shown in red boxes, predictions in green. Black points inside predictions represent pseudo point clouds, while red points indicate real ones.}
\label{fig:vis_moti}
\end{figure*}

\begin{figure}[htpb]
\centering
\includegraphics[scale=0.4]{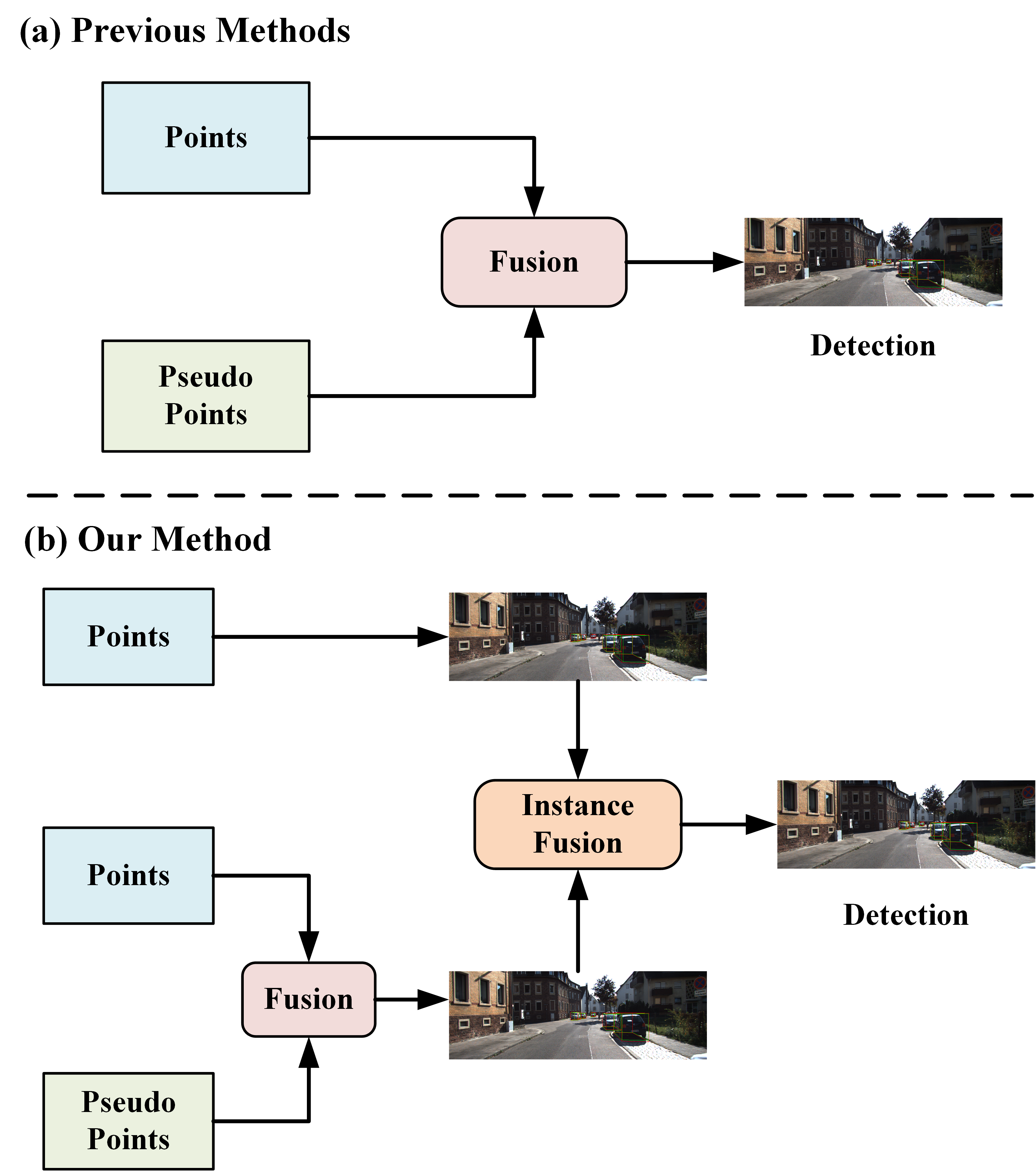}
\caption{\textbf{Comparison of different fusion paradigms.} Previous approaches rely on a symmetric fusion strategy. In contrast, we adopt a LiDAR-dominant two-stage refinement scheme, which integrates instance-level outputs from multiple stages.}
\label{fig:teaser}
\end{figure}

However, LiDAR-based methods are hindered by the sparsity of point clouds, resulting in weaker detection performance for distant or structurally ambiguous objects, as illustrated by blue boxes in Figure~\ref{fig:vis_moti} (a) and (b). To address this limitation, multi-sensor fusion approaches~\cite{chen2017multi, qi2018frustum, vora2020pointpainting, huang2020epnet, chen2022focal, wu2022sparse, li2023logonet, wu2023virtual, mo2024sparse} have attempted to incorporate RGB cameras into 3D perception pipelines. Unlike point clouds, images can provide rich semantic information and are densely distributed on a 2D pixel plane. Given the heterogeneity between images and point clouds, some methods~\cite{wu2022sparse, wu2023virtual, mo2024sparse} perform depth estimation on images and project them into 3D space to generate pseudo point clouds. As illustrated in Figure~\ref{fig:teaser} (a), pseudo point clouds are treated as another type of point cloud and fused equally with real point clouds to generate bounding boxes. Nevertheless, since projecting images into 3D space is an ill-posed problem, the genrated "fake" point clouds inevitably contain noise. In Figure~\ref{fig:vis_moti} (b), pseudo points with uncertain depth and semantics within the purple box cause the detector to misclassify background regions as vehicles. In contrast, LiDAR alleviates false detections through its precise short-range localization capability.

Considering both the advantages and limitations of each sensor, along with the requirement for distance prediction in 3D detection, LiDAR is well-suited to be the primary sensor, while the camera should serve as an auxiliary modality. In light of the above observations and analysis, we present a simple and effective LiDAR-Camera fusion framework, termed LDRFusion, which enables pseudo point clouds and real point clouds to complement each other more effectively. As shown in Figure~\ref{fig:teaser} (b), we adopt a two-stage asymmetric cascade optimization strategy. The first stage of the framework exclusively processes real point clouds to generate initial bounding boxes. In the subsequent stage, both types of point clouds are fed into the network to mitigate the sparsity issue. During inference, the detection results from different stages are integrated through a weighted fusion strategy. This progressive refinement paradigm fully leverages the inherent characteristics of both types of point clouds to improve both recall and precision of the model.

Moreover, the capacity to extract local features from each type of point clouds also influences recognition performance. Due to the modality gap between LiDAR and cameras, designing an appropriate feature encoder for pseudo points remains a problem. Given that existing methods neglect the fine-grained variations among points, we devise a hierarchical pseudo point residual encoding (HPR) module that jointly models positional and feature residuals of pseudo point clouds. This design captures local structural and contextual relationships while maintaining computational efficiency.

Our contributions can be summarized as follows:
\begin{itemize}

\item We propose a straightforward LiDAR-dominant refinement method to fully exploit the complementary feature information between real point clouds and pseudo point clouds.
\item We introduce an efficient feature extractor based on residuals, which enhances the contextual representation ability of each pseudo point cloud.
\item Extensive experiments on the KITTI dataset demonstrate that our method effectively enhances the detection performance of the model without introducing significant computational overhead.
\end{itemize}


\section{Related works}

\subsection{LiDAR-based 3D Detector}
LiDAR-based 3D detectors generate three-dimensional bounding boxes by directly processing raw point cloud data. VoxelNet~\cite{zhou2018voxelnet} utilizes the Voxel Feature Encoder (VFE) to extract features from discretized point clouds. Building upon this, SECOND~\cite{yan2018second} introduces sparse convolution operations to bypass computations on empty voxels, thereby achieving significant speedup without compromising detection accuracy. SA-SSD~\cite{he2020structure} integrates 3D convolutional operations with raw point cloud representations through an auxiliary point-wise supervision network to enhance structural awareness. SE-SSD~\cite{zheng2021se} proposes a distillation paradigm employing paired teacher-student networks to fully leverage soft and hard targets. The aforementioned methods belong to single-stage detection frameworks, which offer advantages in simplicity and inference speed. However, such approaches inevitably lose critical geometric information during point cloud feature downsampling. Consequently, some methods introduce a refinement stage to optimize detection box localization and classification accuracy. PointRCNN~\cite{shi2019pointrcnn} adopts a two-stage architecture based on point representations. Part-A2~\cite{shi2020points} employs Part-Aware and Part-Aggregation methods to predict and aggregate part distribution information respectively. Voxel R-CNN~\cite{deng2021voxel} introduces a Voxel Pooling strategy to extract RoI features, achieving a balance between accuracy and speed. Graph R-CNN~\cite{yang2022graph} constructs local subgraphs for RoIs and iteratively optimizes detection performance. However, LiDAR sensors inherently suffer from point cloud sparsity at long ranges due to their physical measurement principles.

\subsection{Multimodal-based 3D Detector}
Recent approaches have achieved improved detection performance by incorporating image data into the detection pipeline. MV3D~\cite{chen2017multi} and AVOD~\cite{ku2018joint} explore multi-sensor fusion through multi-view aggregation approaches. PointPainting~\cite{vora2020pointpainting} and PointAugmenting~\cite{wang2021pointaugmenting} augment point clouds with pixel-level information using semantic segmentation. \cite{liang2022bevfusion, liu2023bevfusion} project both images and point clouds onto a unified BEV plane to achieve feature alignment. Additionally, some methods enrich raw point clouds by lifting dense image pixels into 3D space. MVP~\cite{yin2021multimodal} utilizes instance segmentation and LiDAR-projected points to perform depth estimation for masked pixels. SFD~\cite{wu2022sparse} employs the depth completion network to generate dense pseudo point clouds, then uses multi-modality Pooling and 3D-GAF to achieve RoI feature fusion. VirConv~\cite{wu2023virtual} applies random dropout to pseudo point clouds for efficiency improvement and utilizes NR-Conv to mitigate noise. MSMDFusion~\cite{jiao2023msmdfusion} lifts pixels to voxel space through MDU-Projection and fuses multi-granularity voxel features using GMA-Conv. SQD~\cite{mo2024sparse} proposes a point-query-pseudo strategy that enables rapid downsampling of pseudo point clouds.

\section{Methods}

\subsection{Overall Framework}
\begin{figure*}[htpb]
\centering
\includegraphics[width=0.7\linewidth]{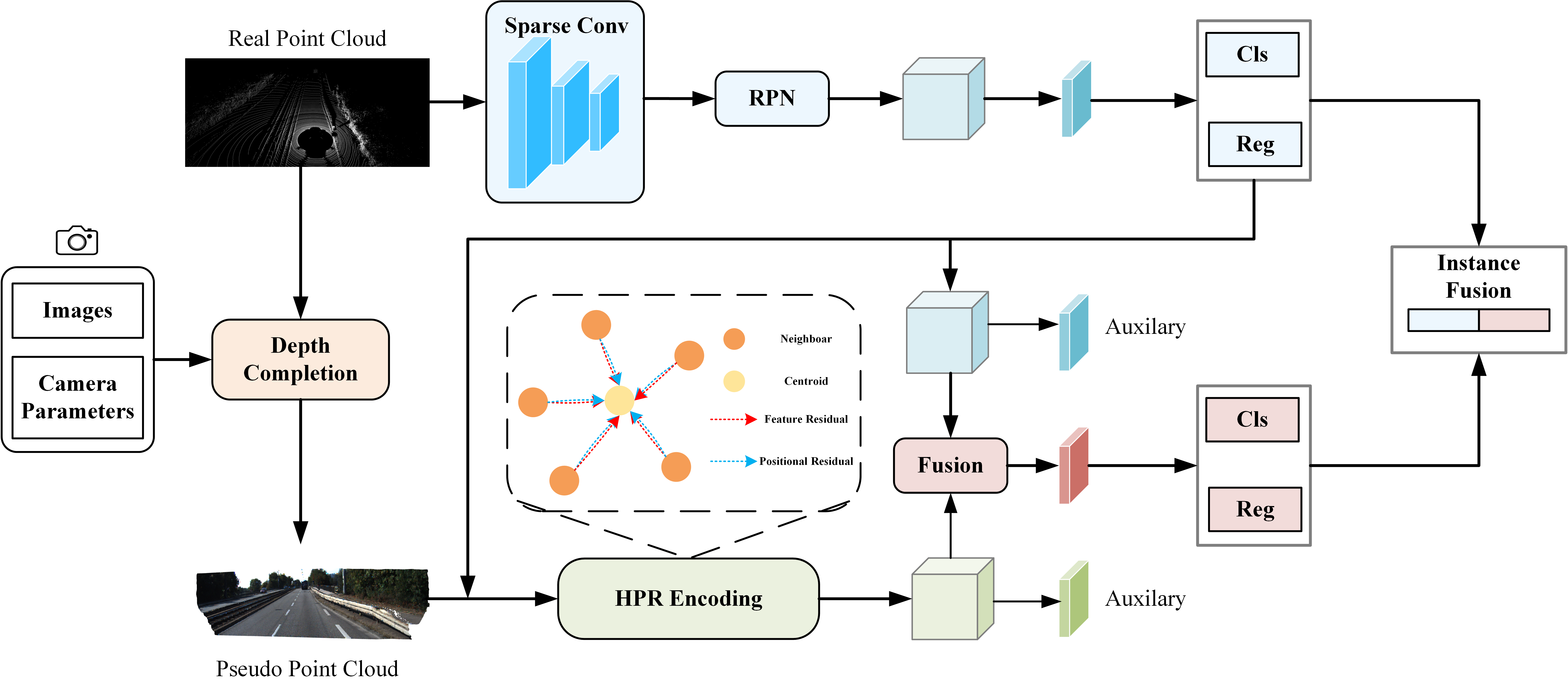}
\caption{\textbf{Overall strucure of LDRFusion.} Our proposed framework primarily focuses on box refinement through a two-stage optimization process. In the first stage, the original point clouds are exclusively utilized, while the second stage integrates dense semantic information from pseudo point clouds to generate precise detection results. When incorporating auxiliary pseudo point clouds, we employ HPR to effectively extract local structural features from the pseudo point data.}
\label{fig:framework}
\end{figure*}

The overall network architecture is illustrated in Figure~\ref{fig:framework}. In Section~\ref{sub32}, we will describe the refinement strategy. In Section~\ref{sub33}, we will demonstrate the design of HPR module. Finally, in Section~\ref{sub34}, we will define the loss function.

\subsection{LiDAR-Dominant Cascade Refinement}
\label{sub32}

To ensure that LiDAR maintains a dominant role in multi-sensor fusion and leverages its precise localization capabilities, we adopt a cascade architecture consisting of two stages. \\
\subsubsection{First Stage Refinement}

In the first stage, the network takes only LiDAR as input. Consistent with conventional methods, the point clouds are processed through voxel-based feature extraction to obtain spatial features $S^L$. Subsequently, a RPN generates a set of RoIs $O^{L}=\{o^{L}_{1}, ..., o^{L}_{N}\}$ in the 3D space. Voxel pooling operation~\cite{deng2021voxel} is then applied to extract voxel features $F^{L}=\{f^{L}_{1}, ..., f^{L}_{N}\}$ within each RoI. These features are fed into regression $R^L$ and classification heads $C^L$ to generate prediction scores and bounding boxes. To further maintain consistency between predicted boxes and their corresponding confidence scores, inspired by ~\cite{he2020structure}, we utilize the RPN to output additional classification maps $\mathcal{X}$ and employ Part-Sensitive Warping (PSW) to obtain auxiliary classification scores, which are averaged with $c_i$ to yield the final confidence score.

\begin{gather}
    f^{L}_i = Pooling(S^L, o^{L}_i) \\
    r^{L}_i = R^L(f^{L}_i), c^{L}_i = C^L(f^{L}_i) + PSW(\mathcal{X}, b^{L}_i) \\
    d^{L}_i = (r^{L}_i, c^{L}_i), i \in \{1, ..., N\}
\end{gather}
where $N$ represents the number of proposals, $b^{L}_i$ are sampling points generated from the predicted bounding boxes. \\
\subsubsection{Second Stage Refinement}

The second stage takes both LiDAR and cameras as input. For a given image $I \in R^{h \times w \times 3}$, depth completion is performed using the network $\varphi$ to obtain a dense depth map. 
Pseudo point clouds $Q$ are then generated using the camera's intrinsic parameter and extrinsic parameter.

A cascaded architecture is then employed to progressively refine the detection outputs from the initial stage. In this optimization phase, the detection instances $D^L$ generated in the first stage are utilized as multi-modal RoIs $O^{M}=\{o^{M}_{1}, ..., o^{M}_{N}\}$ for subsequent processing. For each $o^{m}_{i}$, we perform RoI cropping to extract the encompassed pseudo point clouds within its boundaries. Subsequently, to better capture both semantic and structural information from the pseudo point clouds, we employ the HPR module to extract high-dimensional features $S^C$. Similar to the fisrt stage, pooling is performed on the point clouds from the LiDAR branch and the pseudo point clouds from the camera branch to obtain RoI features $f^{L'}_{i}$ and $f^{C}_{i}$. After that, a fusion network is employed to achieve RoI-level fusion of the two modalities. Eventually, the fused feature $f^{M}_{i}$ is then fed into the detection and regression heads to obtain the detection results $D_{M}=\{d^{M}_{1}, ..., d^{M}_{N}\}$ for the second stage.

\begin{gather}
    o^{M}_i = d^{L}_i \\
    S^C = HPR(Crop(Q, O^M)) \\
    f^{C}_{i} = Pooling(S^C, o^{M}_i), f^{L'}_{i} = Pooling(S^L, o^{M}_i) \\
    f^{M}_{i} = Fusion(f^{C}_{i}, f^{L'}_{i}) \\
    r^{M}_{i} = R^{M}(f^{M}_{i}), c^{M}_{i} = C^{M}(f^{M}_{i}) \\
    d^{M}_{i} = (r^{M}_{i}, c^{M}_{i}), i \in \{1, ..., N\}
\end{gather}

To fully exploit the complementary advantages across refinement stages with heterogeneous point cloud input, during inference, the model weights the instance-level predictions generated at each stage to obtain the final prediction results D.

\begin{equation}
    D = \alpha * D^L + (1 - \alpha) * D^{M}
\end{equation}
where the hyperparameter $\alpha$ is set to 0.5 in our framework.

\subsection{Hierarchical Pseudo Point Residual Encoding}
\label{sub33}
Dense pseudo point clouds are spatially constrained through cropping operations. To further enhance feature extraction in regions of interest, we design a HPR module that focuses on encoding endogenous features and establishing relationships between the pseudo point cloud and their neighbors.

For the j-th pseudo point cloud $q_{i,j}$ within the i-th RoI, it contains nine dimensional information: $(x_{i,j}, y_{i,j}, z_{i,j}, r_{i,j}, g_{i,j}, b_{i,j}, u_{i,j}, v_{i,j}, d_{i,j})$. Here, $(u_{i,j}, v_{i,j})$ represent the image pixel coordinate, $(r_{i,j}, g_{i,j}, b_{i,j})$ denote the corresponding pixel values from $I(u_{i,j}, v_{i,j})$, $(x_{i,j}, y_{i,j}, z_{i,j})$ indicate the spatial coordinates and $d_{i,j}$ is obtained by calculating the L2 norm of its 3D coordinates. For efficient neighborhood querying, we identify K neighboring points by searching for pseudo points corresponding to pixels $(u_{i,j} + \delta u_{i, j, k}, v_{i,j} + \delta v_{i, j, k})$ around $(u_{i,j}, v_{i,j})$, where $\delta u_{i, j, k}\in\{-n, +n\}$ and $\delta v_{i, j, k} \in \{-n, +n\}$ define the search range. 

This approach enables efficient retrieval of the point cloud set $Q_{i,j}=\{q_{i, j, 0}, q_{i, j, 1}, ..., q_{i, j, K}\}$ with $q_{i,j}$ as the centroid, where $q_{i, j, 0}=q_{i,j}$. Inspired by point cloud representation learning methods~\cite{qiu2021pnp, phan2018dgcnn}, we propose a feature and positional residual encoding mechanism for relation construction. High-dimensional point cloud feature differencing aims to eliminate shared attributes between point clouds in that accentuating distinctive characteristics can enable the network to learn discriminative point-wise representations. In detail, for the k-th point $q_{i, j, k}$ in $Q_{i,j}$, $(x_{i, j, k}, y_{i, j, k}, z_{i, j, k}, u_{i, j, k}, v_{i, j, k}, r_{i, j, k})$ are set as initial attributes to encode as $s^{0}_{i, j, k}$ via a Multi-Layer Perceptron (MLP) to encapsulate both 3D structural and 2D semantic information. This representation is then progressively refined through multi-step iterative aggregation. At step $t+1$, we construct feature vector by concatenating the feature residual $(s^{t}_{i, j, k} - s^{t}_{i, j, 0})$ with respect to the centroid and its own feature $s^{t}_{i, j, k}$. Simultaneously, the positional residual $(p_{i, j, k} - p_{i, j, 0})$ is processed through an MLP to reweight features. In contrast to the implicit difference encoding in feature residuals, positional residuals explicitly model joint 3D-2D spatial relationships between pseudo point clouds, serving as a basis for assigning weights to different neighboring points. $s^{t+1}_{i, j, k}$ are then updated through the dot product of the positional residual and feature vector. Due to the lack of rotational invariance in pseudo point clouds after introducing RGB features, we replace max pooling operations used in~\cite{qi2017pointnet, phan2018dgcnn} with a MLP to aggregate each neighboring point, yielding the $s^{t+1}_{i, j}$. Eventually, the features from each iteration are concatenated to construct $s_{i, j}$, which is used as the final feature representation of $Q_{i, j}$.

\begin{gather}
    s^{t+1}_{i, j, k} = \mathrm{Concat}(s^{t}_{i, j, k} - s^{t}_{i, j, 0}, s^{t}_{i, j, k}) * \mathcal{M}_{\theta}^{t}(p_{i, j, k} - p_{i, j, 0}) \\
    s^{t+1}_{i, j} = \mathcal{M}_{\gamma}^{t}(s^{t+1}_{i, j, 0}, ..., s^{t+1}_{i, j, K}) \\
    s_{i, j} = \mathrm{Concat}(s^{1}_{i, j}, ..., s^{t+1}_{i, j})
\end{gather}
where ${\mathcal{M}_{\theta}^{t}}$ and ${\mathcal{M}_{\gamma}^{t}}$ represent the MLP networks.\\

\subsection{Training Losses}
\label{sub34}
The overall loss function follows configurations of~\cite{deng2021voxel, wu2022sparse}, consisting of RPN and RoI loss. The RoI loss consists of two components: (1) the first-stage LiDAR-based detection loss $\mathcal{L}^{L}$. (2) the second-stage losses including pseudo point cloud auxiliary supervision loss $\mathcal{L}^{L}_{aux}$, raw point cloud auxiliary supervision loss $\mathcal{L}^{C}_{aux}$, and multimodal fusion loss $\mathcal{L}^{M}$. All losses mentioned above employ Focal loss as classification loss and regression part uses Smooth L1 loss. Besides, $\mathcal{L}^{M}$ incorporates GIoU~\cite{rezatofighi2019generalized} loss for enhanced bounding box regression.

\begin{gather}
    \mathcal{L} = \mathcal{L_{RPN}} + \lambda_1\mathcal{L}^{L} + \lambda_2(\mathcal{L}^{L}_{aux} + \mathcal{L}^{C}_{aux}) + \lambda_3 \mathcal{L}^{M}
\end{gather}
where the loss weighting coefficients are set as $\lambda_1 = 1.0$, $\lambda_2 = 0.5$, and $\lambda_3 = 1.0$ to balance the contributions of different loss components.

\section{Experiments}
\subsection{Dataset}

This work conducts experiments on the KITTI benmark~\cite{geiger2013vision}, which provides RGB cameras and 64-line LiDAR data. The dataset consists of 7481 training frames and 7518 test frames. The training data can be further divided into a training set of 3712 frames and a validation set of 3769 frames. The evaluated categories include cars, pedestrians, and cyclists, with respective IoU thresholds configured at 0.7, 0.7, and 0.5. Based on the occlusion level and size of the ground truth bounding boxes, the detection difficulty is categorized into easy, moderate, and hard. The evaluation metrics used are 3D averaged precision (AP) calculated by 40 recall points ($R_{40}$) and 11 recall points ($R_{11}$). Additionally, evaluation is also carried out under the bird’s eye view (BEV) perspective.

We also evaluate our method on the nuScenes dataset~\cite{caesar2020nuscenes}, which offers large-scale multimodal sensor data with 3D annotations, and adopt the official evaluation metrics including mean Average Precision (mAP) and nuScenes Detection Score (NDS).

\subsection{Implementation Details}

All experiments are implemented based on the open-source 3D object detection framework OpenPCDet. Our work follows the training configuration of SFD~\cite{wu2022sparse} and employs its data augmentation strategies, including random flipping, rotation, scaling, local noise addition, and synchronized ground truth sampling specifically designed for pseudo point clouds. The network is trained on two 3090 GPUs with a batch size of 2. We use the pre-trained depth completion network TWISE~\cite{imran2021depth} to generate pseudo point clouds.

\subsection{Main Results}
\subsubsection{Detection Performance}
\begin{table*}[htpb]
\centering
\caption{Performance for car detection on the KITTI test set (\%).}
\begin{tabular}{ccllll}
\toprule
\multicolumn{1}{c}{\multirow{2}{*}{Method}} & \multicolumn{1}{c}{\multirow{2}{*}{Reference}}  & \multicolumn{4}{c}{3D Car AP ($R_{40}$)} \\ \cline{3-6} 
\multicolumn{1}{c}{} & \multicolumn{1}{c}{} & \multicolumn{1}{c}{Easy} & \multicolumn{1}{c}{Mod.} & \multicolumn{1}{c}{Hard} & \multicolumn{1}{c}{mAP}\\
\midrule
Voxel R-CNN~\cite{deng2021voxel} & AAAI 2021 & 90.90 & 81.62 & 77.06 & 83.19\\
CasA~\cite{wu2022casa} & TGRS 2022  & 91.58	& 83.06 & 80.08 & 84.91 \\
GraR-Vo~\cite{yang2022graph} & ECCV 2022  & 91.89 & 83.27 & 77.78 & 84.31 \\
SFD~\cite{wu2022sparse} & CVPR 2022  & 91.73 & 84.76 & 77.92 
& 84.80\\
Focals Conv~\cite{chen2022focal} & CVPR 2022  & 90.55 & 82.28 & 77.59 & 83.47\\
GLENet~\cite{zhang2023glenet} & IJCV 2023  & 91.67 & 83.23 & 78.43 & 83.47 \\
3Onet~\cite{hoang20233onet} & IEEE SENS J 2023  & \textbf{92.03} & 85.47 & 78.64 & 85.38 \\
LoGoNet~\cite{li2023logonet} & CVPR 2023  & 91.80 & 85.06 & \textbf{80.74} & 85.87 \\
PVT-SSD~\cite{yang2023pvt} & CVPR 2023  & 90.65 & 82.29 & 76.85 & 83.26 \\
TED~\cite{wu2023transformation} & AAAI 2023  & 91.61 & 85.28 & 80.68 & 85.86 \\
SQD~\cite{mo2024sparse} & ACMM 2024  & 91.58 & 81.82 & 79.07 & 84.16 \\
SparseDet~\cite{liu2024sparsedet} & TGRS 2024  & 90.79 & 81.17 & 78.11 & 83.36 \\
SLBEVFusion~\cite{nie2025investigating} & Neural Computing 2025  & 88.63 & 80.67 & 73.47 & 80.92 \\
SSLFusion~\cite{ding2025sslfusion} & AAAI 2025  & 91.40 & 84.40 & 80.00 & 85.27 \\
ViKIENet~\cite{yu2025viki} & CVPR 2025  & 91.79 & 84.96 & 80.20 & 85.65 \\
LDRFusion (Ours) &  & 91.92 & \textbf{85.47} & 80.43 & \textbf{85.94} \\ 
\bottomrule
\end{tabular}
\label{tab:testcar}
\end{table*}

\begin{table}[htpb]
\centering
\caption{Performance for car detection on the KITTI validation set (\%).}
\resizebox{\columnwidth}{!}{
\begin{tabular}{cllllll}
\toprule
\multicolumn{1}{c}{\multirow{2}{*}{Method}} & \multicolumn{3}{c}{3D Car AP ($R_{40}$)} & \multicolumn{3}{c}{3D Car AP ($R_{11}$)} \\ \cline{2-7} 
\multicolumn{1}{c}{} & \multicolumn{1}{c}{Easy} & \multicolumn{1}{c}{Mod.} & \multicolumn{1}{c}{Hard} & \multicolumn{1}{c}{Easy} & \multicolumn{1}{c}{Mod.} & \multicolumn{1}{c}{Hard} \\
\midrule
Voxel R-CNN~\cite{deng2021voxel} & 92.38 & 85.29 & 82.86 & 89.41 & 84.52 & 78.93 \\
SFD~\cite{wu2022sparse} & 95.47 & 88.56 & 85.74 & 89.74 & 87.12 & 85.20 \\
SQD~\cite{mo2024sparse} & 95.39 & 85.40 & 85.02 & 89.72 & 83.84 & 83.98 \\
LDRFusion(Ours) & \textbf{96.00} & \textbf{89.06}& \textbf{86.47} & \textbf{90.24} & \textbf{87.72} & \textbf{86.21} \\ 
\bottomrule
\end{tabular}}
\label{tab:valcar}
\end{table}

In the Table~\ref{tab:testcar} comparative results demonstrate substantial performance gains of our method over the LiDAR-based Voxel R-CNN~\cite{deng2021voxel} across all evaluation metrics. In comparison to the multimodal baseline SFD~\cite{wu2022sparse}, it achieves consistent improvements of 0.19\%, 0.71\%, and 2.51\% on easy, moderate, and hard difficulty levels respectively. The above results indicate that integrating single-modal and multi-modal information during the optimization stage can effectively improve the model's detection accuracy. Furthermore, the proposed method achieves outstanding performance in 3D and BEV mAP compared to other advanced detectors, demonstrating its superior localization capability. Table~\ref{tab:valcar} further confirms the performance advantages through improvements in both 3D AP ($R_{40}$) and AP ($R_{11}$) on the KITTI validation set under the setting where training is conducted on the car category alone.

Table~\ref{tab:valthreeclass} comprehensively evaluates our method's multi-category detection capability. Notably, under joint training across object classes, our approach achieves an overall mAP of 78.71\%, marking a 0.92\% improvement over the previously best-performing method. In particular, for rigid objects such as cars, it surpasses prior approaches across all difficulty levels. For smaller classes like pedestrians and cyclists, where depth estimation is inherently more challenging compared to larger vehicles, our approach does not achieve the highest mAP. Nevertheless, the refinement strategy based on accurate point clouds strengthens the ability to identify these challenging targets, attaining second-best performance.
\begin{table*}[htbp]
\centering
\caption{Performance for multi-class detection on the KITTI validation set (\%).}
\resizebox{\textwidth}{!}{
\begin{tabular}{ccccccccccccccc}
\toprule
\multicolumn{1}{c}{\multirow{2}{*}{Method}} & \multicolumn{4}{c}{3D Car AP ($R_{40}$)} & \multicolumn{4}{c}{3D Pedestrian AP ($R_{40}$)} & \multicolumn{4}{c}{3D Cyclist AP ($R_{40}$)} & \multicolumn{1}{c}{\multirow{2}{*}{3D mAP}}\\ \cline{2-13} 
\multicolumn{1}{c}{} & \multicolumn{1}{c}{Easy} & \multicolumn{1}{c}{Mod.} & \multicolumn{1}{c}{Hard} & \multicolumn{1}{c}{mAP} & \multicolumn{1}{c}{Easy} & \multicolumn{1}{c}{Mod.} & \multicolumn{1}{c}{Hard} & \multicolumn{1}{c}{mAP} & \multicolumn{1}{c}{Easy} & \multicolumn{1}{c}{Mod.} & \multicolumn{1}{c}{Hard} & \multicolumn{1}{c}{mAP}\\
\midrule
SECOND~\cite{yan2018second} & 88.61 & 78.62 & 77.22 & 81.48 & 56.55 & 52.98 & 47.73 & 52.42 & 80.58 & 67.15 & 63.10 & 70.28 & 68.06 \\
PointPillars~\cite{lang2019pointpillars} & 86.46 & 77.28 & 74.65 & 79.46 & 57.75 & 52.29 & 47.90 & 52.65 & 80.05 & 62.68 & 59.70 & 67.48 & 66.53 \\
PDV~\cite{hu2022point} & 92.56 & 85.29 & 83.05 & 86.97 & 66.90 & 60.80 & 55.85 & 61.18 & 92.72 & 74.23 & 69.60 & 78.85 & 75.67 \\
F-PointNet~\cite{qi2018frustum} & 83.76 & 70.92 & 63.65 & 72.78 & 70.00 & 61.32 & 53.59 & 61.64 & 77.15 & 56.49 & 53.37 & 62.34 & 65.58 \\
EPNet~\cite{huang2020epnet} & 88.76 & 78.65 & 78.32 & 81.91 & 66.74 & 59.29 & 54.82 & 60.28 & 83.88 & 65.60 & 62.70 & 70.69 & 70.96 \\
VFF~\cite{li2022voxel} & 92.31 & 85.51 & 82.92 & 86.91 & 73.26 & 65.11 & 60.03 & 66.13 & 89.40 & 73.12 & 69.86 & 77.46 & 76.84 \\
CasA~\cite{wu2022casa} & 93.21 & 86.37 & 83.93 & 87.84 & \textbf{73.95} & \textbf{66.62} & 59.97 & \textbf{66.85} & \textbf{92.78} & 73.94 & 69.37 & 78.70 & 77.79 \\
LogoNet~\cite{li2023logonet} & 92.04 & 85.04 & 84.31 & 87.13 & 70.20 & 63.72 & 59.46 & 64.46 & 91.74 & 75.35 & \textbf{72.42} & \textbf{79.84} & 77.14 \\
LDRFusion(Ours) & \textbf{95.86} & \textbf{88.77}& \textbf{86.39} & \textbf{90.34} & 73.67 & 66.12 & \textbf{60.06} & 66.62 & 91.32 & \textbf{75.42} & 70.75 & 79.16 & \textbf{78.71}\\ 
\bottomrule
\end{tabular}}
\label{tab:valthreeclass}
\end{table*}

To further verify the generalization capability of our model, we conduct experiments on the nuScenes dataset. As shown in Table~\ref{tab:nusc_res}, our method surpasses the two-stage MVP~\cite{yin2021multimodal} approach with improvements of 0.8\% in mAP and 0.3\% in NDS. These results demonstrate that our approach serves as an effective mechanism that can be applied to other pseudo point based methods.

\begin{table}[htbp]
\centering
\caption{Performance for detection on the nuScenes validation set (\%).}
\begin{tabular}{ccc}
\toprule
Method & mAP & NDS \\
\midrule
CenterPoint~\cite{yin2021center} & 56.4 & 64.8 \\
MVP~\cite{yin2021multimodal} & 66.0 & 69.9 \\
MVP + 2 stage~\cite{yin2021multimodal} & 67.0 & 70.7 \\
MVP + 2 stage + LPRFusion & \textbf{67.8} & \textbf{71.0} \\
\bottomrule
\end{tabular}
\label{tab:nusc_res}
\end{table}

\subsubsection{Inference Speed}

Table~\ref{tab:inf_time} shows the inference speed of different models. Compared to several high-performance detectors~\cite{hoang20233onet, wu2023transformation} that incorporate complex data augmentation, our method achieves a relatively fast speed of 10 FPS. Moreover, while surpassing SFD~\cite{wu2022sparse} in performance, our approach shows only a marginal decrease of 0.2 FPS.

\begin{table}[htpb]
\centering
\caption{Inference speed of different methods.}
\resizebox{\columnwidth}{!}{
\begin{tabular}{cccc}
\hline
\textbf{LDRFusion(Ours)} & \textbf{SFD}~\cite{wu2022sparse} & \textbf{3ONet}~\cite{hoang20233onet} & \textbf{TSSTDet} \cite{hoang2024tsstdet} \\
10.0 FPS & 10.2 FPS & 6.5 FPS & 7.7 FPS \\ \hline
\textbf{TED}~\cite{wu2023transformation} & \textbf{CasA}~\cite{wu2022casa} & \textbf{Voxel R-CNN}~\cite{deng2021voxel} &  \textbf{SQD}~\cite{mo2024sparse} \\
11.1 FPS & 11.6 FPS & 21.08 FPS & 13.5 FPS \\ \hline
\end{tabular}}
\label{tab:inf_time}
\end{table}

\subsection{Ablation Study}

\subsubsection{Effective of Key Components}

Table~\ref{tab:ablation} presents the car detection results of individual modules in our proposed method on the KITTI validation set, where (a) corresponds to the baseline~\cite{wu2022sparse} used for comparative evaluation. The setting (b) demonstrates that our multi-stage refinement network brings steady gains of 0.32\%, 0.35\%, and 0.64\% AP on the easy, moderate, and hard difficulty levels respectively compared to the baseline. As indicated in (c) of the table, the enhanced HPR feature encoding module achieves a 0.26\% mAP increase when integrated into the baseline. Further, entry (d) shows that incorporating the HPR module during the second-stage refinement of configuration (b) yields additional performance gains across all difficulty levels.

\begin{table}[htpb]
\centering
\caption{Effects of key components (\%).}
\begin{tabular}{ccccccc}
\toprule
\multirow{2}{*}{} & \multirow{2}{*}{\textit{Refine.}} & \multirow{2}{*}{\textit{HPR}} & \multicolumn{4}{c}{3D Car AP ($R_{40}$)} 
\\ \cline{4-6}  \cline{7-7}
 & & & Easy & Mod. & Hard. & mAP\\
\midrule
(a) & & & 95.47 & 88.56 & 85.74 & 89.92 \\
(b)  & \checkmark & & 95.73 & 88.93 & 86.39 & 90.35\\
(c)  &  & \checkmark & 95.37 & 89.05 & 86.13 & 90.18 \\ 
(d)  & \checkmark & \checkmark & 96.00 & 89.06 & 86.47 & 90.51 \\ 
\bottomrule
\end{tabular}
\label{tab:ablation}
\end{table}

\subsubsection{Different Refine Policies}
\label{subsub432}

Table~\ref{tab:diffrefine} provides a detailed comparison of various refinement strategies. We first investigate how distinct modality configurations influence performance across refinement stages. In setting (\uppercase\expandafter{\romannumeral1}), only LiDAR is used, while in setting (\uppercase\expandafter{\romannumeral2}), both real point clouds and pseudo point clouds are used throughout all stages. The comparison results reveal that incorporating pseudo point clouds as auxiliary input leads to improvements across multiple metrics. Nevertheless, compared to (\uppercase\expandafter{\romannumeral4}), introducing pseudo point clouds in both stages results in a performance decline. This suggests that emphasizing the contribution of real point clouds during the refinement stage is beneficial for the network to perceive more precise spatial information. Furthermore, we explore a alternative fusion strategy. Method (\uppercase\expandafter{\romannumeral4}) adopts feature-level fusion as in CasA~\cite{wu2022casa}, and the results show that it achieves performance comparable to instance-level fusion. However, feature fusion entails increased computational overhead and memory consumption, which slows down both training and inference. Therefore, instance-level fusion emerges as a more efficient and preferable solution.


\begin{table*}[htpb]
\centering
\caption{Different refine policies (\%).}
\begin{tabular}{cccccccccccc}
\toprule
\multirow{2}{*}{} & \multirow{2}{*}{First Stage} & \multirow{2}{*}{Second Stage} & \multirow{2}{*}{Fusion Strategy} & \multicolumn{3}{c}{BEV Car AP ($R_{40}$)}  & \multicolumn{3}{c}{3D Car AP ($R_{40}$)} 
\\ \cline{5-10}
 & & & & \multicolumn{1}{c}{Easy} & \multicolumn{1}{c}{Mod.} & \multicolumn{1}{c}{Hard} & \multicolumn{1}{c}{Easy} & \multicolumn{1}{c}{Mod.} & \multicolumn{1}{c}{Hard} \\
\midrule
(\uppercase\expandafter{\romannumeral1}) & Real & Real & Instance & 96.34 & 91.84 & 91.30 & 93.19 & 85.78 & 83.33 \\
(\uppercase\expandafter{\romannumeral2}) & Real + Pseudo & Real + Pseudo & Instance & 96.19 & 92.05 & 91.53 & 95.26 & 88.26 & 85.73 \\
(\uppercase\expandafter{\romannumeral3}) & Real & Real + Pseudo & Feature & 96.52 & 92.41 & 91.89 & 95.76 & 88.84 & 86.32 \\ 
(\uppercase\expandafter{\romannumeral4}) & Real  & Real + Pseudo & Instance & 96.51 & 92.66 & 92.12 & 95.67 & 89.12 & 86.30 \\ 
\bottomrule
\end{tabular}
\label{tab:diffrefine}
\end{table*}

\section{Conclusion}
This paper introduces a novel 3D object detection framework, LDRFusion, which incorporates a LiDAR-centric refinement strategy and a dedicated feature encoder for pseudo point clouds. The proposed approach effectively leverages the strengths of multi-modal inputs while maintaining computational efficiency. Extensive experiments demonstrate the efficacy of our method. Future work may explore integrating this framework with other advanced detectors to further enhance detection performance.

\footnotesize

\normalsize
\bibliography{reference}


\footnotesize



\end{document}